\documentclass[final]{cvpr}

\usepackage{times}
\usepackage{epsfig}
\usepackage{graphicx}
\usepackage{amsmath}
\usepackage{amssymb}

\usepackage{wrapfig}
\usepackage{comment}
\usepackage{multirow}
\usepackage{subfigure}
\usepackage[dvipsnames]{xcolor}
\usepackage[ruled,vlined]{algorithm2e}

\usepackage[pagebackref=true,breaklinks=true,colorlinks,bookmarks=false]{hyperref}

\def\cvprPaperID{7264} 
\def\confYear{CVPR 2021}

\usepackage{nopageno}

\begin{document}


\title{Track, Check, Repeat: An EM Approach to Unsupervised Tracking}


\makeatletter
\newcommand{\printfnsymbol}[1]{%
  \textsuperscript{\@fnsymbol{#1}}%
}
\makeatother

\author{%
Adam W. Harley 
\quad\quad\quad Yiming Zuo\thanks{Equal contribution.}
\quad\quad\quad Jing Wen\printfnsymbol{1}
\quad\quad\quad Ayush Mangal\\
\quad\quad\quad Shubhankar Potdar
\quad\quad\quad Ritwick Chaudhry
\quad\quad\quad Katerina Fragkiadaki \vspace{.5em} \\
Carnegie Mellon University \\
{\tt\small \{aharley, yzuo, jingwen2, smpotdar, rchaudhr, katef\}@cs.cmu.edu, amangal@cs.iitr.ac.in}
\vspace{-.5em}
}

\maketitle

\newcommand\adam[1]{\textcolor{magenta}{#1}}
\newcommand\todo[1]{\textcolor{red}{#1}}
\newcommand\red[1]{\textcolor{red}{#1}}
\newcommand\gist[1]{\textcolor{cyan}{#1}}

\newcommand{\M}{\mathcal{M}} 
\newcommand{\R}{\mathcal{R}} 
\newcommand{\Mt}{\mathcal{M}^{(t)}} 
\newcommand{\Mz}{\mathcal{M}^{(0)}} 
\newcommand{\Mo}{\mathcal{M}^{(1)}} 
\newcommand{\Mi}{\mathcal{M}^{(i)}} 
\newcommand{\Mj}{\mathcal{M}^{(j)}} 
\newcommand{\Mto}{\mathcal{M}^{(t+1)}} 

\newcommand{\Tt}{\mathcal{T}^{(t)}} 
\newcommand{\Tz}{\mathcal{T}^{(0)}} 
\newcommand{\To}{\mathcal{T}^{(1)}} 

\newcommand{\I}{I} 
\newcommand{\D}{D} 

\newcommand{\loss}{\mathcal{L}}

\newcommand{\Ot}{{O}^{(t)}}
\newcommand{\Oz}{{O}^{(1)}}
\newcommand{\Ct}{{O}^{(t)}}
\newcommand{\Cht}{\hat{{O}}^{(t)}}
\newcommand{\Iht}{\hat{{I}}^{(t)}}

\begin{abstract}
We propose an unsupervised method for detecting and tracking moving objects in 3D, in unlabelled RGB-D videos. The method begins with classic handcrafted techniques for segmenting objects using motion cues: we estimate optical flow and camera motion, and conservatively segment regions that appear to be moving independently of the background. Treating these initial segments as pseudo-labels, we learn an ensemble of appearance-based 2D and 3D detectors, under heavy data augmentation. We use this ensemble to detect new instances of the ``moving'' type, even if they are not moving, and add these as new pseudo-labels. Our method is an expectation-maximization algorithm, where in the expectation step we fire all modules and look for agreement among them, and in the maximization step we re-train the modules to improve this agreement. The constraint of ensemble agreement helps combat contamination of the generated pseudo-labels (during the E step), and data augmentation helps the modules generalize to yet-unlabelled data (during the M step). We compare against existing unsupervised object discovery and tracking methods, using challenging videos from CATER and KITTI, and show strong improvements over the state-of-the-art.


\end{abstract}

\vspace{-1em}
\section{Introduction} \label{sec:intro}


Humans can detect moving objects and delineate their approximate extent~\cite{spelke1982perceptual,craton1996development}, without ever having been supplied boxes or segmentation masks as supervision. 
Where does this remarkable ability come from? Psychology literature points to a variety of perceptual grouping cues, which make some regions look more object-like than others \cite{goldstein2009perceiving}. These types of objectness cues have long been known in computer vision literature \cite{alexe2010object}, yet this domain knowledge has not yet led to powerful self-supervised object detectors. 

Work on integrating perceptual grouping cues into computer vision models stretches back decades \cite{sarkar1993perceptual}, and still likely serves as inspiration for many of the design decisions in modern computer vision architectures related to attention, segmentation, and tracking. 
Much of the current work on recognition and tracking is fully-supervised, and relies on vast pools of human-provided annotations. On the unsupervised side, a variety of deep learning-based methods have been proposed, which hinge on reconstruction objectives and part-centric, object-centric, or scene-centric bottlenecks in the architecture \cite{kulkarni2019unsupervised,burgess2019monet}. These methods are rapidly advancing, but so far only on toy worlds, made up of simple 2D or 3D shapes against simple backgrounds -- a far cry from the complexity tackled in older works, based on perceptual grouping (e.g., \cite{Fragkiadaki:topology}).

Classic methods of object discovery, such as center-surround saliency in color or flow \cite{alexe2012measuring}, are known to be brittle, but they need not be discarded entirely. In this paper, we propose to mine and exploit the admittedly rare success scenarios of these models, to bootstrap the learning of something more general. We hypothesize that if the successful vs. unsuccessful runs of the classic algorithms can be readily identified with automatic techniques, then we can self-supervise a learning-based module to mimic and outperform the traditional methods. This is a kind of knowledge distillation \cite{hinton2015distilling}, from traditional models to deep ones.

We propose an optimization algorithm for learning detectors of moving objects, based on expectation maximization~\cite{nigam2000text}. We begin with a motion-based handcrafted detector, tuned to be very conservative (low recall, high precision). We then convert each object proposal into thousands of training examples for learning-based 2D and 3D detectors, by randomizing properties like color, scale, and orientation. This forces the learned models to generalize, and allows recall to expand. 
We then use the ensemble of models to obtain new high-confidence estimates \textbf{(E step)}, repeat the optimization \textbf{(M step)}, and iterate. Our method outperforms not only the traditional methods, which only work under specific conditions, but also the current learning-based methods, which only work in toy environments. We demonstrate success in a popular synthetic environment where recent deep models have already been deployed (CLEVR/CATER \cite{johnson2016clevr,girdhar2020cater}), and also on the real-world urban scenes benchmark (KITTI \cite{Geiger2013IJRR}), where the existing learned models fall flat.

Our main contribution is not in any particular component, but rather in their combination. We demonstrate that by exploiting the successful outcomes of traditional methods for moving object segmentation, we can train a learning-based method to detect and track objects in a target domain, without requiring any annotations. 

\section{Related  Work} \label{sec:related}


\paragraph{Object discovery}
Many recent works have proposed deep neural networks for object discovery from RGB videos. These models typically have an object-centric bottleneck, and are tasked with a reconstruction objective. 
MONet \cite{burgess2019monet}, Slot attention \cite{locatello2020object}, IODINE \cite{greff2019multi} , SCALOR \cite{Jiang*2020SCALOR:}, AIR \cite{NIPS2016_52947e0a}, and AlignNet \cite{creswell2020alignnet} fall under this category. These methods have been successful in a variety of simple domains, but have not yet been tested on real-world videos. 
In this paper we evaluate whether these models are able to perform well under the complexities of real-world imagery. 


\vspace{-1em}\paragraph{Ensemble methods}

Using ensembles is a well-known way to improve overall performance of an algorithm. Assuming that each member of the ensemble is prone to different types of errors, the combination of them is likely to make fewer errors than any individual component \cite{dietterich2000ensemble}. Ensembling is also the key idea behind knowledge distillation, where knowledge gets transferred from cumbersome models to simpler ones \cite{hinton2015distilling,Radosavovic_2018_CVPR}. A typical modern setup is to make up the ensemble out of multiple copies of a neural network, which are trained from different random initializations or using different partitions of the data \cite{allen2020towards}. In our case, the ensemble is more diverse: it is made up of components which solve different tasks, but which can still be checked against one another for consistency. For example, we learn a 2D pixel labeller which operates on RGB images, and a 3D object detector which operates on voxelized pointclouds; when the 3D detections are projected into the image, we expect them to land on ``object'' pixels. 

\vspace{-1em}\paragraph{Never ending learning}

We take inspiration from the methods described in ``never ending learning'' literature \cite{10.5555/2898607.2898816,chen2013neil}, where the goal is to learn an ever-increasing set of concepts over the course of an infinite training loop. 
We follow the ``macro-vision'' philosophy described by Chen et al.~\cite{chen2013neil}, and build dataset-level knowledge (i.e., detectors) by leveraging a small number of ``easy'' examples from a dataset, rather than attempt to understand every data point. 

\vspace{-1em}\paragraph{Structure-from-Motion/SLAM}
Early works on structure from motion (SfM) \cite{Tomasi:1992:SMI:144398.144403,10.1109/ICCV.1995.466815} set the ambitious goal of extracting unscaled 3D scene pointclouds and camera trajectories from 2D pixel trajectories, exploiting the reduced rank of the trajectory matrix under rigid motions. Unfortunately, these methods are often confined to very simple videos, due to their difficulty handling camera motion degeneracies, non rigid object motion, or frequent occlusions, which cause 2D trajectories to be short in length. 
Simultaneous Localization And Mapping (SLAM) methods  optimize the camera poses in every frame as well as the 3D coordinates of points in the scene online and often in real time, assuming a calibrated setup  (i.e., knowing camera intrinsics) \cite{schoeps14ismar,kerl13iros}. These methods are sensitive to measurement noise and the difficulties of multi-view correspondence, but produce accurate reconstructions when assumptions on the sensor and scene setup are met. Dynamic objects are typically treated as outliers \cite{keller2013real,yang2020teaser}, or are actively detected with the help of optical flow \cite{cheng2019improving} or pre-trained appearance cues \cite{barnes2018driven}. 
Our method exploits the occasional successes of a flow-based egomotion estimation method as a starting point to learn about the static vs. moving parts of scenes. 

\vspace{-1em}\paragraph{Moving object segmentation} 
Early approaches attempted motion segmentation completely unsupervised  by integrating motion information over time through 2D pixel trajectories \cite{springerlink:10.1007/978-3-642-15555-0_21,OB11}. Recent works instead focus on learning to segment 2D objects in videos, supervised by annotated video benchmarks \cite{bhat2020learning,Hu_2019_CVPR,8099855,li2020fast,Fragkiadaki_2015_CVPR}. 
Here, we segment a sparse set of objects using motion cues, then learn to segment in 2D and 3D using appearance, without annotations.

\vspace{-1em}\paragraph{Learning from augmentations}
The state of the art approach in self-supervised learning of 1D visual representations (i.e., vectors describing images) relies on training the features to be invariant to random augmentations, such as color jittering and random cropping \cite{he2019momentum,chen2020simple}. Some tracking approaches use data augmentation at test time, to fine-tune the tracker with diverse variations of a specified target 
\cite{DBLP:journals/corr/KhorevaBIBS17}. Interestingly, the most important factor seems to be high diversity, rather than realism, even for practical robotics applications \cite{tobin2017domain}. 
Our work takes inspiration from these methods, to upgrade \textit{self-generated} annotations into diverse datasets through data augmentation.

\begin{figure*}[t!]
\centering{
 \includegraphics[width=1.0\linewidth]{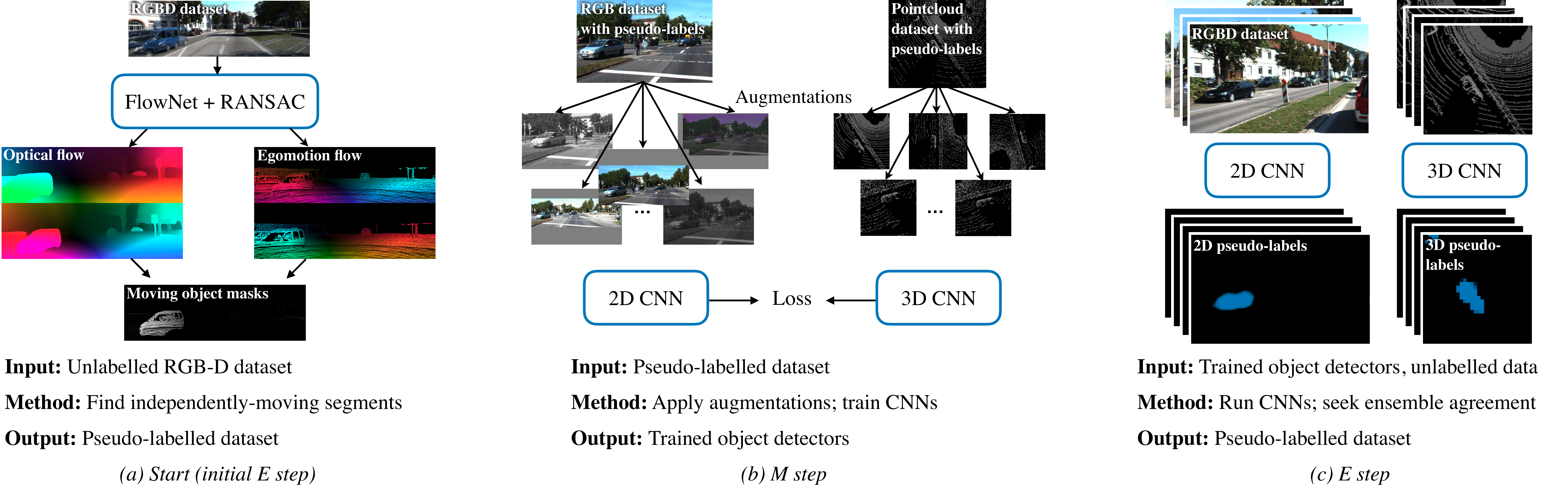}
 }
 \vspace{-0.5em}
 \caption{ 
 \textbf{An EM approach to unsupervised tracking.} We present an expectation-maximization (EM) method, which takes RGBD videos as input, and produces object detectors and trackers as output. \textbf{(a)} We begin with a handcrafted E step, which uses optical flow and egomotion to segment a small number of objects moving independently from the background. \textbf{(b)} Next, as an M step, we treat these segmented objects as pseudo-labels, and train 2D and 3D convolutional nets to detect these objects under heavy data augmentation. \textbf{(c)} We then use the learned detectors as an ensemble to re-label the data (E step), and loop. 
 }
 \label{fig:overview}
\end{figure*}

\section{Track, Check, Repeat} \label{sec:model}

\subsection{Setup and overview}
Our method takes as input a video with RGB and depth (either a depthmap or a pointcloud) and camera intrinsics, and produces as output a 3D detector and 3D tracker for moving rigid objects in the video. 

We treat this data as a test set, in the sense that we do not use any annotations. In the current literature, most machine learning methods have a training phase and a test phase, where the model is ``frozen'' when the test phase arrives. Our method instead attempts to optimize its parameters for the test domain, using ``free'' supervision that it automatically generates (without any human intervention).

The optimization operates in ``rounds''. The first round leverages optical flow and cycle-consistency constraints to discover a small number of clearly-moving objects in the videos. The most confident object proposals are upgraded into pseudolabels for training appearance-based object detectors. The second round leverages optical flow and the new objectness detectors to find more high-confidence proposals, which again lead to additional training. In the final round we use the detectors as trackers, and use these to generate a library of trajectories, capturing a motion prior for objects in the domain. 

A critical piece in each stage is the ``check'', which decides whether or not to promote an estimate into a pseudolabel for the next round. We now describe each piece of the method, along with its corresponding check.

\subsection{Optical flow estimation} \label{sec:flow}

Optical flow indicates a 2D motion field that corresponds the pixels of a pair of images. We use flow (in combination with other cues) as a signal for objectness, and also as a submodule of egomotion estimation. 

We use an off-the-shelf pre-trained convolutional optical flow network \cite{raftnet} to estimate flow. The pre-training involves dense supervision with synthetic frame pairs, but we note that optical flow can be learned unsupervised on real data \cite{back_to_basics:2016,selflow}. In our experiments, we found that the pre-trained model generalized well, and self-supervised finetuning did not improve accuracy further. 

For our purposes, it is important to avoid relying on flow vectors which are likely to be incorrect. To do this, we check for forward-backward consistency \cite{wu2007situ,pan2009recurrent,sethi1987finding}. We first generate the forward flow $f_{0 \to 1}$ and the backward flow $f_{1 \to 0}$ with the flow network, and then warp the backward flow into the coordinates of the first image,
\begin{equation}
    \hat{f}_{1 \to 0}= \text{warp}(f_{1 \to  0} ; f_{0 \to 1} ).
\end{equation}
This uses a bilinear warping function, which takes as arguments an image to warp (in this case $f_{1 \to 0}$), and a flow field to warp with (in this case $f_{0 \to 1}$). We then threshold on the size of the discrepancy between flow fields, using a threshold that is sensitive to the flow magnitude:
\begin{equation}
   || f_{0 \to  1}+ \hat{f}_{1 \to 0} ||_2 <\alpha_1 \left( || f_{0 \to  1} ||_2 + ||\hat{f}_{1 \to  0} ||_2 \right) + \alpha_2.
\end{equation}
Note that the discrepancy term uses the \textit{addition} of the flow fields rather than the difference, since the flows point in opposite directions. We use $\alpha_1 = 0.01$ and $\alpha_2 = 0.1$ in our experiments.

\subsection{Egomotion estimation} \label{sec:egomotion}

Egomotion is the rigid motion of the camera (i.e., transformation of poses) across a pair of frames. Estimating egomotion allows us to better estimate which pixels are moving due to the camera's motion, and which are moving independently. Pixels moving independently are a strong cue for objectness. 

We begin by ``upgrading'' the cycle-consistent 2D flows into a sparse 3D pointcloud flow. 
To do this, we first obtain sparse 2D depth maps, by projecting the pointclouds into pixel coordinates.
We then check each flow vector to see if it starts and ends at pixels with depth measurements. We then use the flows and corresponding depths to un-project the flow into a 3D motion field. 

We then use RANSAC to estimate the 6-degrees-of-freedom rigid motion that explains the maximal number of point flows. RANSAC is intended to be robust to outliers, but the answer returned is often catastrophically wrong, due either to correspondence errors or moving objects. 

The critical third step is to ``check'' the RANSAC output with a freely-available signal. The inlier count itself is such a signal, but this demands carefully tuning the threshold for inlier counting. Instead, we enforce cycle-consistency, similar to flow. We estimate two rigid motions with RANSAC: once using the forward flow, and once using the backward flow (which delivers an estimate of the inverse transform, or backward egomotion). We then measure the inconsistency of these results, by applying the forward and backward motion to the \textit{same} pointcloud, and measuring the maximum alignment error:

\begin{equation}
x_{0}' = T_{1 \to 0}^{bw} T_{0 \to 1}^{fw}(x_0)
\end{equation}
\begin{equation}
\text{error} = \max_n ( || x_0'  - x_0 || ),
\end{equation}
where $T_{0 \to 1}^{fw}$ denotes the rotation and translation computed from forward flow, which carries the pointcloud from timestep $0$ to timestep $1$, $T_{1 \to 0}^{bw}$ is the backward counterpart, and $x_0$ denotes the pointcloud from timestep $0$. 


If the maximum displacement across the entire pointcloud is below a threshold (set to 0.25 meters), then we treat the estimate as ``correct''. In practice we find that this occurs about 80\% of the time in the KITTI dataset. 

\begin{figure}[t!]
\centering{
 \includegraphics[width=1.0\linewidth]{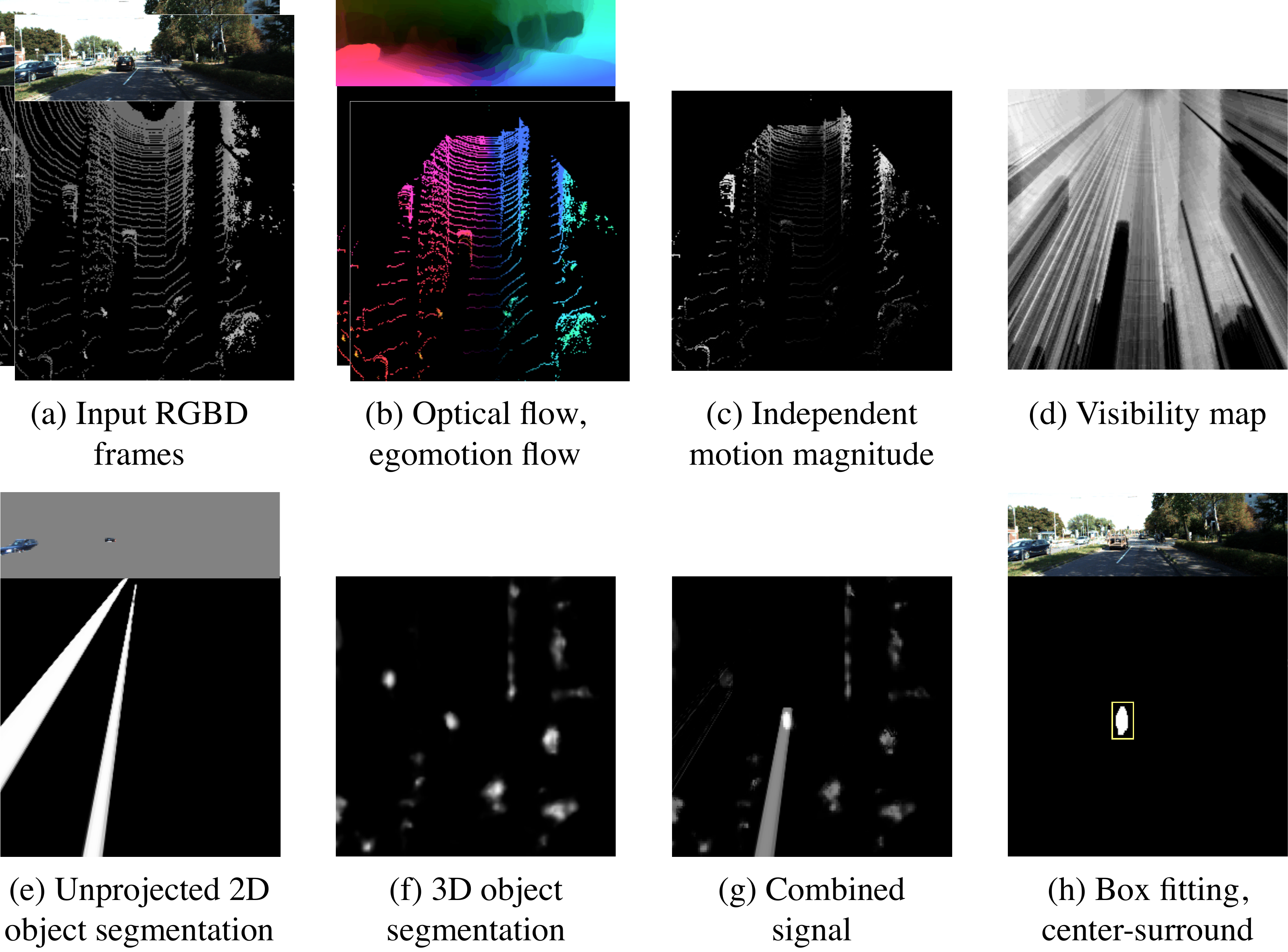}
 }
 \vspace{-0.5em}
 \caption{ 
 \textbf{Intermediate outputs of the \textit{Expectation} step in our algorithm.} Given an input video (a), we have multiple sources of evidence: using optical flow and egomotion flow (b), we compute independent motion magnitude (c); using the pointcloud we compute visible area (d), using the RGB image we estimate 2D segmentation (e), and using the pointcloud we estimate 3D segmentation (f). Each signal can be error-prone, but combining them (g) gives us high confidence pseudo-labels (h).
 }
 \label{fig:pseudogen}
\end{figure}

On these successful runs, we apply the egomotion to the pointcloud to create another 3D flow field (in addition to the one produced by upgrading the optical flow to 3D), and we subtract these to obtain the camera-independent motion field. Independently moving objects produce high-magnitude regions in the egomotion-stabilized motion field, which is an excellent cue for objectness. Examples of this are shown in Figure~\ref{fig:overview}-a and Figure~\ref{fig:pseudogen}-c: note that although real objects are highlighted by this field, some spurious background elements are highlighted also. 

In the first ``round'' of optimization, we proceed directly from this stage to pseudo-label generation. Using the pseudo-labels, we train the parameters of two object detectors, described next.


\subsection{2D objectness segmentation} 

This module takes an RGB image as input, and produces a binary map as output. The intent of the binary map is to estimate the likelihood that a pixel belongs to the ``moving object'' class. This module transfers knowledge from the motion-based estimators into the domain of appearance, since it learns to mimic pseudolabels that were generated from motion alone. This is an important aspect of the overall model, since it allows us to identify objects of the ``moving'' type even when they are stationary. 

We use a 50-layer ResNet \cite{he2016deep} with a feature pyramid \cite{lin2017feature} as the architecture, and train the last layer with a logistic loss against sparse pseudo-ground-truth:
\begin{equation}
\loss^\text{seg} = \sum \hat{m} \log(1 + \exp(- \hat{s} \cdot s) ),
\end{equation}
where $m$ is a mask indicating where the supervision is valid. We experimented with and without ImageNet pretraining for the ResNet, and found that the pretrained version converges more quickly but does not perform very differently. 

In training this module with sparse labels, it is critical to add heavy augmentations to the input, so that it does not simply memorize a mapping from the happenstance appearance to the sparse objectness labels. We use random color jittering, random translation and scaling, and random synthetic occlusions. 

\subsection{3D object detection}

This module takes as input a voxelized pointcloud (computed from the depth map and intrinsics), and estimates object proposals in 3D. We have experimented with producing oriented 3D bounding boxes, or 3D voxel segmentations, with similar results. 

We use a 3D U-Net-style convolutional encoder \cite{ronneberger2015u}, and a CenterNet-style detection head \cite{duan2019centernet}. The head produces a set of heatmaps, which encode objectness (in 1 channel), 3D size (in 3 channels), 3D subvoxel offset (in 3 channels), and orientation along the vertical axis (encoded as a categorical distribution over 16 channels). For implementation details we refer the reader to the supplementary file, and the original 2D CenterNet paper \cite{duan2019centernet}. 

In training this module, we find that randomized translation and orientation (when creating the voxelized input) are critical for learning an even distribution over possible object orientations. Additionally, we apply dropout in the voxel inputs, and we create partial occlusion augmentations by randomly erasing a randomly-sized area near the object pseudo-label in image space, along with the 3D points that project into that area. 

\subsection{Short-range tracking}

To relocate a detected object over short time periods, we use two simple techniques: hungarian matching with IoU scores, and cross correlation with a rigid template \cite{matthews2004template}. We find that the IoU method is sufficient in CATER, where the motions are relatively slow. In KITTI, due to the fast motions of the objects and the additional camera motion, we find that cross-correlation is more effective. We do this using the features provided by the backbone of the object detector. We simply create a template by encoding a crop around the object, and then use this template for 3D cross correlation against features produced in nearby frames. We find that this is a surprisingly effective tracker despite not handling rotations, likely because the objects do not undergo large rotations under short timescales. To track for longer periods and across occlusions, we make use of motion priors represented in a library of previously-observed motions, described next. 

\subsection{Long-range tracking, with trajectory libraries}

To track objects over longer time periods, we build and use a library of motion trajectories, to act as a motion prior. 
We build the library out of the successful outcomes of short-range tracker, which typically correspond to ``easy'' tracking cases, such as close-range objects will full visibility. The key insight here is that a motion prior built from ``good visibility'' tracklets is just as applicable to ``poor visibility tracklets'', since visibility is not a factor in objects' motion. 

To verify tracklets and upgrade them into library entries, we check if they agree with the per-timestep cues, provided by flow, 2D segmentation, 3D object detection, and a visibility map computed by raycasting on the pointcloud. Specifically, we ask that a tracklet (1) obey the flow field, and (2) travel through area that is either object-like or invisible. For flow agreement, we simply project the 3D object motion to 2D and measure the inconsistency with the 2D flow in the projected region. To ensure that the trajectory travels through object-like territory, we create a spatiotemporal volume of objectness/visiblity cues, and trilinearly sample in that volume at each timestep along the trajectory. Each temporal slice of the volume is given by:
\begin{equation}
    p = \max(\textrm{unproj}(s) \cdot o + (1.0 - v), 1),
\end{equation}
where $\textrm{unproj}(s)$ is the 2D segmentation map unprojected to 3D (Figure~\ref{fig:pseudogen}-d), $o$ is the 3D heatmap delivered by the object detector (Figure~\ref{fig:pseudogen}-f), and $v$ is the visibility map computed through raycasting (Figure~\ref{fig:pseudogen}-d). In other words, we require that both the 2D and 3D objectness signals agree on the object's presence, or that the visibility indicates the object is in an occluded area. To evaluate a trajectory's likelihood, we simply take the mean of its values in the spatiotemporal volume, and we set a stringent threshold (0.99) to prevent erroneous tracklets from entering the library. 

Once the library is built, we use it to link detections across partial and full occlusions (where flow-based and correlation-based tracking fails). Specifically, we orient the library to the initial motion of an object, and then evaluate the likelihood of all paths in the library, via the cost volume. This is similar to a recent approach for motion planning for self-driving vehicles \cite{zeng2019end}, but here the set of possible trajectories is generated from data rather than handcrafted.

\subsection{Pseudo-label generation}

Pseudo-label generation is what takes the model from one round of optimization to the next. The intent is to select the object proposals that are likely to be correct, and treat them as ground truth for training future modules. 

We take inspiration from never-ending learning architectures \cite{mitchell2018never}, which promote an estimate into a label only if (i) at least one module produces exceedingly-high confidence in the estimate, or (ii) multiple modules have reasonably-high confidence in the estimate.

The 2D and 3D modules directly produce objectness confidences, but the motion cues need to be converted into an objectness cue. Our strategy is inspired by classic literature on motion saliency \cite{itti2000saliency}: (1) compute the magnitude of the egomotion-stabilized 3D motion field, (2) threshold it at a value (to mark regions with motion larger than some speed), (3) find connected components in that binary map (to obtain discrete regions), and (4) evaluate the center-surround saliency of each region. Specifically, we compute histograms of the motion inside the region and in the surrounding shell, compute the chi-square distance between the distributions, and threshold on this 
value \cite{alexe2012measuring}:
\begin{equation}
    cs(\theta) = \chi^2 (h(\textrm{cen}_\theta( \Delta x)),  h(\textrm{surr}_\theta( \Delta x))),
\end{equation}
where $\theta$ denotes the region being evaluated, $\textrm{cen}_\theta$ and $\textrm{surr}_\theta$ select points within and surrounding the region, $h$ computes a histogram, and $\Delta x$ denotes the egomotion-stabilized 3D motion field. 

When the trained objectness detectors are available (i.e., on rounds after the first), we convert the egomotion-stabilized motion field into a heatmap with $\exp(-\lambda ||\Delta x||)$, and add this heatmap to the ones produced by the 2D and 3D objectness estimators. We then proceed with thresholding, connected components, and box fitting as normal. The only difference is that we use a threshold that demands multiple modules to agree: since each module produces confidences in $[0,1]$, setting the threshold to any value above $2$ enforces this constraint.

\subsection{Connection to EM}

The mathematical connection to Expectation-Maximization (EM) is not perfect, because our model is not generative, but the mapping is quite close. Following the terminology of Nigam et al.~\cite{nigam2000text}, we have: 

\vspace{-0.5em}\paragraph{E step: \textit{Use the current classifier to estimate the class membership for each datapoint.} } In our case, use the ensemble of handcrafted and learned modules to estimate the objectness probability for each pixel/point. 

\vspace{-0.5em}\paragraph{M step: \textit{Re-estimate the classifier, using the estimated class labels.}} In our case, re-optimize the parameters of the learned components of the ensemble, using estimates produced by the ensemble as a whole (i.e., where agreement was reached). 

Note that if our classifier were a \textit{single} discriminative model, it would theoretically converge after the first M step; using an \textit{ensemble} of independent modules allows our method to improve over rounds, since it will not converge until all submodules produce the same labelling.

\begin{figure}[t!]
\centering{
 \includegraphics[width=1.0\linewidth]{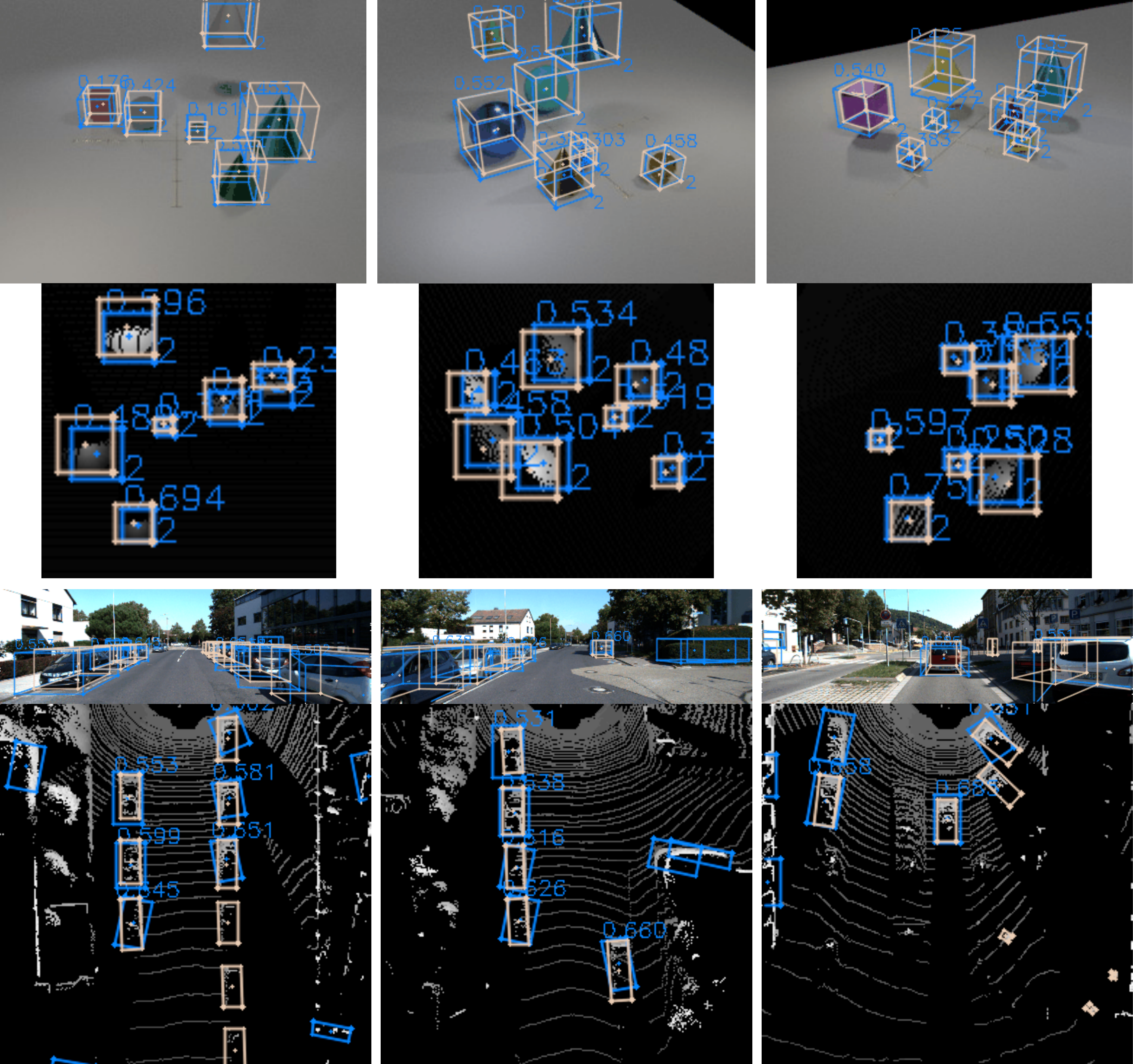}
 }
 \vspace{-0.5em}
 \caption{ 
 \textbf{3D object detections in CATER (top) and KITTI (bottom).} Ground-truth boxes are shown in beige and detection results are shown in blue. IoU scores are marked alongside each box. Results are shown in perspective RGB and bird's-eye view.
 }
 \label{fig:dets}
\end{figure}

\section{Experiments}\label{sec:exp}

We evaluate in the following datasets: 
\begin{enumerate} 
\item Synthetic RGB-D videos of tabletop scenes from CATER \cite{girdhar2020cater}. CATER is a built upon CLEVR \cite{johnson2016clevr}, and it focuses on testing a model's ability to do long-term temporal reasoning. We modified the simulator so that it can generate depth maps in addition to RGB images, but leave all other rendering parameters untouched. The max number of objects is set to 10 to make the scenes as complex as possible. Videos are captured by 6 virtual cameras placed around the scene. 

\item Real RGB-D videos of urban scenes, from the KITTI dataset \cite{Geiger2013IJRR}. This data was collected with a sensor platform mounted on a moving vehicle, with a human driver navigating through a variety of road types in Germany. The data provides multiple images per timestep; we use the ``left'' color camera. The dataset provides depth in the form of LiDAR sweeps synced to the images. 
We use the ``tracking'' subset of KITTI, which includes 3D object labels, and approximate (but relatively inaccurate) egomotion. 
\end{enumerate}
We evaluate the models on their ability to discover objects in 3D, and track objects over time. 

\subsection{Baselines} 

For unsupervised object discovery, we use the following baselines: 

\begin{itemize}
\item \vspace{-0.5em} \textbf{Object-Centric Learning with Slot Attention \cite{locatello2020object}}. This model trains an autoencoder with image reconstruction loss, with soft clustering assignments as a representational bottleneck. An iterative attention-based update mechanism is applied when constructing the clusters.

\item \vspace{-0.5em} \textbf{MONet \cite{burgess2019monet}}. MONet applies a recurrent attention mechanism that tries to explain the scene part by part, with a VAE. 
Since there is no official code released, we re-implemented it. 

\item \vspace{-0.5em} \textbf{Spectral Clustering \cite{brox2010object}}. This method first extracts dense point trajectories using optical flow, then performs object segmentation by computing affinities between trajectories and then clustering.

\item \vspace{-0.5em} \textbf{Discontinuity-Aware Clustering \cite{fragkiadaki2012video}}. This method also begins with dense point trajectories, but uses density discontinuities in the spectral embeddings to improve the segmentation. 
\vspace{-0.5em}

\end{itemize}

For tracking, we use the following baselines:

\begin{itemize}
    \item \vspace{-0.5em} \textbf{Supervised Siamese Network \cite{bertinetto2016fully}}. This is a 3D convolutional net trained as a siamese object tracker. This model produces a feature volume for the object at $t=0$, and produces feature volume for the scene at each timestep, and then locates the object in the wider scene by doing cross correlation at each step. The model is supervised so that the peak of the correlation heatmap is in the correct place. 
    \item \vspace{-0.5em} \textbf{Tracking by Colorizing \cite{vondrick2018tracking}}. This model learns features through an image reconstruction task. The goal is to colorize a grayscale image, by indexing into a source color image with feature dot products. We upgrade this model into a 3D tracker by associating the learned features to the 3D pointcloud instead of RGB values, and by doing RANSAC on the point-wise correspondences to estimate rigid object motions \cite{tracking_emerges_3d}.
\end{itemize}
\vspace{-1em} 
\begin{table*}[t]
\begin{center}
\label{tab:my-table}
\begin{tabular}{cllllllll}
\hline
\multirow{2}{*}{\textbf{Method}} & \multicolumn{1}{c}{\multirow{2}{*}{\textbf{Dataset}}} & \multicolumn{7}{c}{\textbf{mAP@X}} \\
 & \multicolumn{1}{c}{} & \multicolumn{1}{c}{\textbf{0.1}} & \multicolumn{1}{c}{\textbf{0.2}} & \multicolumn{1}{c}{\textbf{0.3}} & \multicolumn{1}{c}{\textbf{0.4}} & \multicolumn{1}{c}{\textbf{0.5}} & \multicolumn{1}{c}{\textbf{0.6}} & \multicolumn{1}{c}{\textbf{0.7}} \\ \hline
\multirow{2}{*}{\begin{tabular}[c]{@{}c@{}}Slot Attention \cite{locatello2020object}\end{tabular}} & CATER (2D) & 0.63 & 0.51 & 0.43 & 0.34 & 0.22 & 0.1 & 0.05 \\
 & KITTI (2D) & 0.07 & 0.03 & 0.01 & 0 & 0 & 0 & 0 \\ \hline
\multirow{2}{*}{MONet \cite{burgess2019monet}} & CATER (2D) & 0.23 & 0.14 & 0.12 & 0.10 & 0.07 & 0.03 & 0.01 \\
 & KITTI (2D) & 0.03 & 0.01 & 0 & 0 & 0 & 0 & 0 \\ \hline
\multirow{2}{*}{\begin{tabular}[c]{@{}c@{}}Spectral Clustering \cite{brox2010object}\end{tabular}} & CATER (2D) & 0.18 & 0.08 & 0.04 & 0.03 & 0.01 & 0 & 0 \\
 & KITTI (2D) & 0.08 & 0.03 & 0.02 & 0.02 & 0.01 & 0 & 0 \\ \hline
\multirow{2}{*}{\begin{tabular}[c]{@{}c@{}}Discontinuity\\ Aware Clustering \cite{fragkiadaki2012video}\end{tabular}} & CATER (2D) & 0.17 & 0.08 & 0.04 & 0.02 & 0.01 & 0.01 & 0 \\
 & KITTI (2D) & 0.08 & 0.04 & 0.03 & 0.01 & 0 & 0 & 0 \\ \hline
\multirow{4}{*}{\begin{tabular}[c]{@{}c@{}}Ours\\ (Round1)\end{tabular}} & CATER (2D) & \textbf{0.98} & 0.97 & \textbf{0.97} & 0.94 & 0.86 & 0.7 & \textbf{0.36} \\
 & KITTI (2D) & \textbf{0.53} & 0.39 & 0.18 & 0.06 & 0.03 & 0.01 & 0.01 \\
 & CATER (BEV) & 0.97 & 0.92 & 0.75 & 0.57 & 0.34 & 0.06 & 0 \\
 & KITTI (BEV) & \textbf{0.46} & \textbf{0.42} & 0.06 & 0 & 0 & 0 & 0 \\ \hline
\multirow{4}{*}{\begin{tabular}[c]{@{}c@{}}Ours\\ (Round2)\end{tabular}} & CATER (2D) & \textbf{0.98} & 0.97 & 0.96 & 0.94 & \textbf{0.88} & 0.69 & 0.33 \\
 & KITTI (2D) & 0.43 & \textbf{0.40} & \textbf{0.39} & 0.33 & 0.30 & 0.22 & 0.10 \\
 & CATER (BEV) & 0.97 & 0.95 & 0.84 & 0.66 & \textbf{0.46} & 0.08 & 0 \\
 & KITTI (BEV) & 0.41 & 0.39 & \textbf{0.35} & 0.31 & 0.28 & 0.11 & 0.02 \\ \hline
\multirow{4}{*}{\begin{tabular}[c]{@{}c@{}}Ours\\ (Round3)\end{tabular}} & CATER (2D) & \textbf{0.98} & \textbf{0.98} & \textbf{0.97} & \textbf{0.95} & \textbf{0.88} & \textbf{0.71} & 0.34 \\
 & KITTI (2D) & 0.43 & \textbf{0.4} & 0.37 & \textbf{0.35} & \textbf{0.33} & \textbf{0.3} & \textbf{0.21} \\
 & CATER (BEV) & \textbf{0.98} & \textbf{0.97} & \textbf{0.9} & \textbf{0.76} & \textbf{0.46} & \textbf{0.1} & \textbf{0.02} \\
 & KITTI (BEV) & 0.4 & 0.38 & \textbf{0.35} & \textbf{0.33} & \textbf{0.31} & \textbf{0.23} & \textbf{0.06} \\ \hline
\end{tabular}%
\end{center}
\caption{\textbf{Object discovery performance, in CATER and KITTI.} Results are reported as mean average precision (mAP) at several IoU threshols. Our method works best in all the metrics reported. 2D means perspective view and BEV means bird's-eye view.}
\end{table*}

\begin{figure}[t]
 \includegraphics[width=\linewidth]{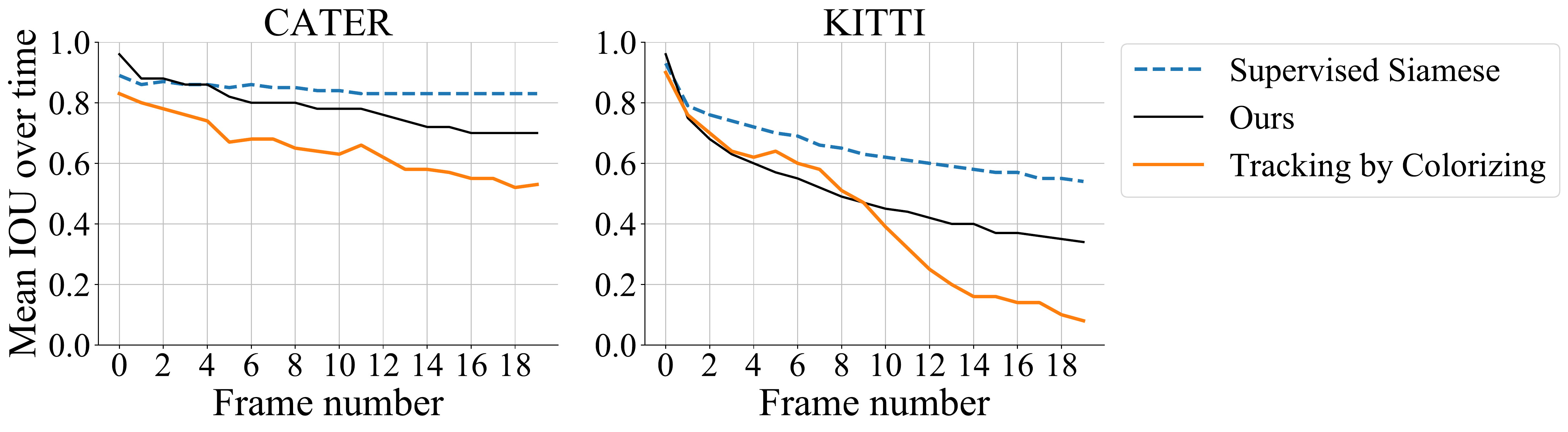}
 \caption{ 
 \textbf{3D object tracking IoU over time, in CATER and KITTI.} Tracking precision necessarily begins near 1.0 because tracking is initialized with a real object box in frame0, and declines over time, more drastically in KITTI than in CATER.}
 \label{fig:tracking_baselines}
\end{figure}


\subsection{Quantitative Results}
\paragraph{Object discovery} Our main evaluation is in Table~\ref{tab:my-table}, where we evaluate the object proposal accuracy in mean Average Precision (mAP) at different IoU thresholds, on both CATER and KITTI. The metrics are collected in a bird's-eye view (BEV) and in 2D projections (2D). We find that our model outperforms the baselines in nearly all metrics. Adding additional rounds of EM improves the precision at the higher IoU thresholds. Interestingly, the learning-based baselines, which achieve state-of-the-art results in synthetic datasets, have near zero accuracy in KITTI. This is probably because certain assumptions in their design are violated in this data (e.g., a static-camera assumption is violated, and there is relatively little self-similarity within object/background regions compared to CATER). See the supplementary file for visualizations of the baselines' outputs. 

\vspace{-1em}\paragraph{Object tracking} Object tracking accuracy (in IoU over time) is shown in Figure~\ref{fig:tracking_baselines}. To evaluate tracking, we initialize the correlation tracker with the bounding box of the object to track. 

As shown in Figure~\ref{fig:tracking_baselines}, the supervised model outperforms the unsupervised ones, especially in CATER (where the data and supervision are perfect), and by a narrower margin in KITTI. Our method can maintain relatively high IoU over long time horizons. The IoU at frame19 is 0.34 in KITTI and 0.7 in CATER. Our model compares favorably to the 3D-upgraded colorization baseline. 

We also input our Round3 KITTI detections to a recent tracking-by-detection method \cite{Weng2020_AB3DMOT_eccvw}, and evaluated with standard multi-object tracking metrics. We obtained sAMOTA: 0.2990; AMOTA: 0.0863; AMOTP: 0.1581. This is encouraging but still far behind the supervised state-of-the-art, which obtains sAMOTA: 0.9328; AMOTA: 0.4543; AMOTP: 0.7741. Our main failure case appears to be missed detections. We also used this tracker to compute alternate results for the IoU-over-time evaluation (cf. Fig.~\ref{fig:tracking_baselines}): this yields a relatively stable line around 0.44 IoU across 20 frames. Performance appears upper-bounded by the detector's mean IoU.

\begin{figure}[t]
 \includegraphics[width=1.0\linewidth]{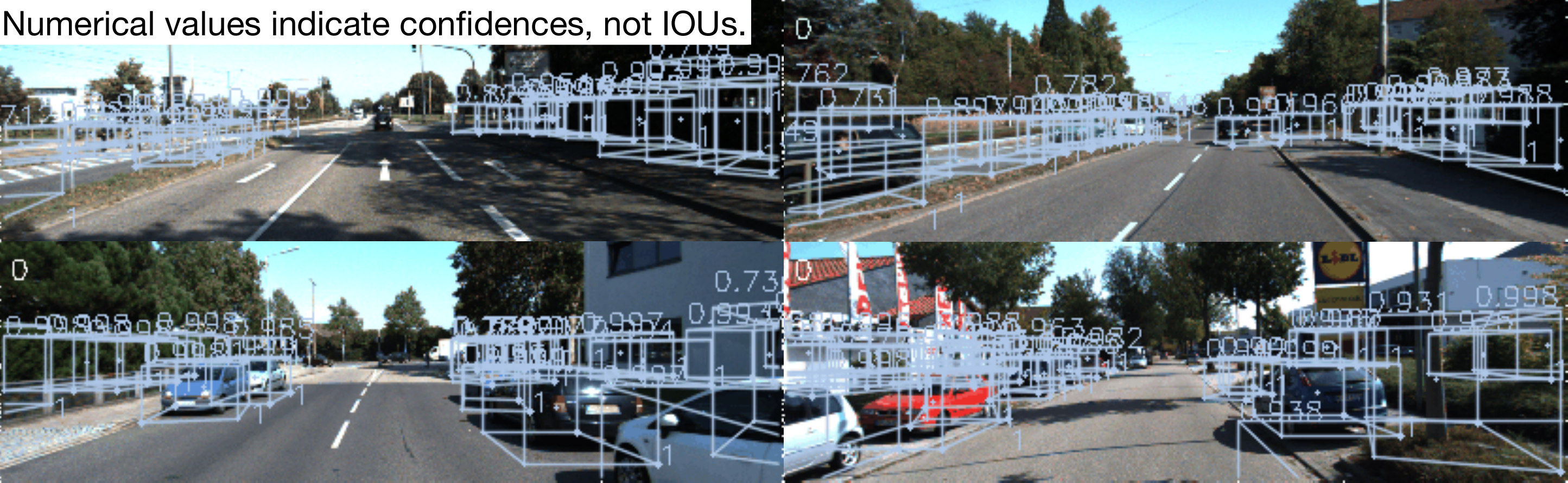}
 \caption{ 
 \textbf{Ablation of ensemble agreement causes divergence.} The model eventually detects objects everywhere.}
 \label{fig:divergence}
\end{figure}

\vspace{-1em}\paragraph{Ablation on ensemble agreement} 
Figure~\ref{fig:divergence} shows what happens when the ensemble agreement check is dropped: the model gradually begins classifying everything as an object (BEV mAP@.5=0.17 on Round2 instead of 0.28). 

\vspace{-1em}\paragraph{Ablation on the trajectory library}  Table~\ref{tab:traj_lib} shows an ablation study on the trajectory library. We report ``Recall'', which we define as the proportion of objects that are successfully tracked by our model from the beginning of the video to the end, where tracking success is defined by an IoU threshold of 0.5. 
We also report ``Precision'', which we define as the proportion of tracklets that begin and end on the same object. 
With the trajectory library, we improve the recall from 53\% to 64\%, while precision drops slightly from 94\% to 91\%. 
Qualitatively we find that the majority of improvement is on partially and fully-occluded objects, where strict appearance-based matching is ambiguous and prone to failure, but where the library is a useful prior. 

We present the remainder of the quantitative results in the supplementary, showing that the tracking performance of our model outperforms the baseline, and showing that ablating components of our model decreases performance.

\begin{table}
\begin{center}
\label{tab:traj_lib}
\begin{tabular}{lll}
\hline
\textbf{Method} & \textbf{Recall} & \textbf{Precision} \\
\hline
Ours, with short-range tracker & 0.53 & 0.94\\
\ldots and trajectory library & 0.64 & 0.91\\
\hline
\end{tabular}
\end{center}
\caption{Ablations of the trajectory library, in CATER.}
\vspace{-1em}
\end{table}

\subsection{Qualitative Results} 
\begin{figure}[t!]
 \includegraphics[width=1.0\linewidth]{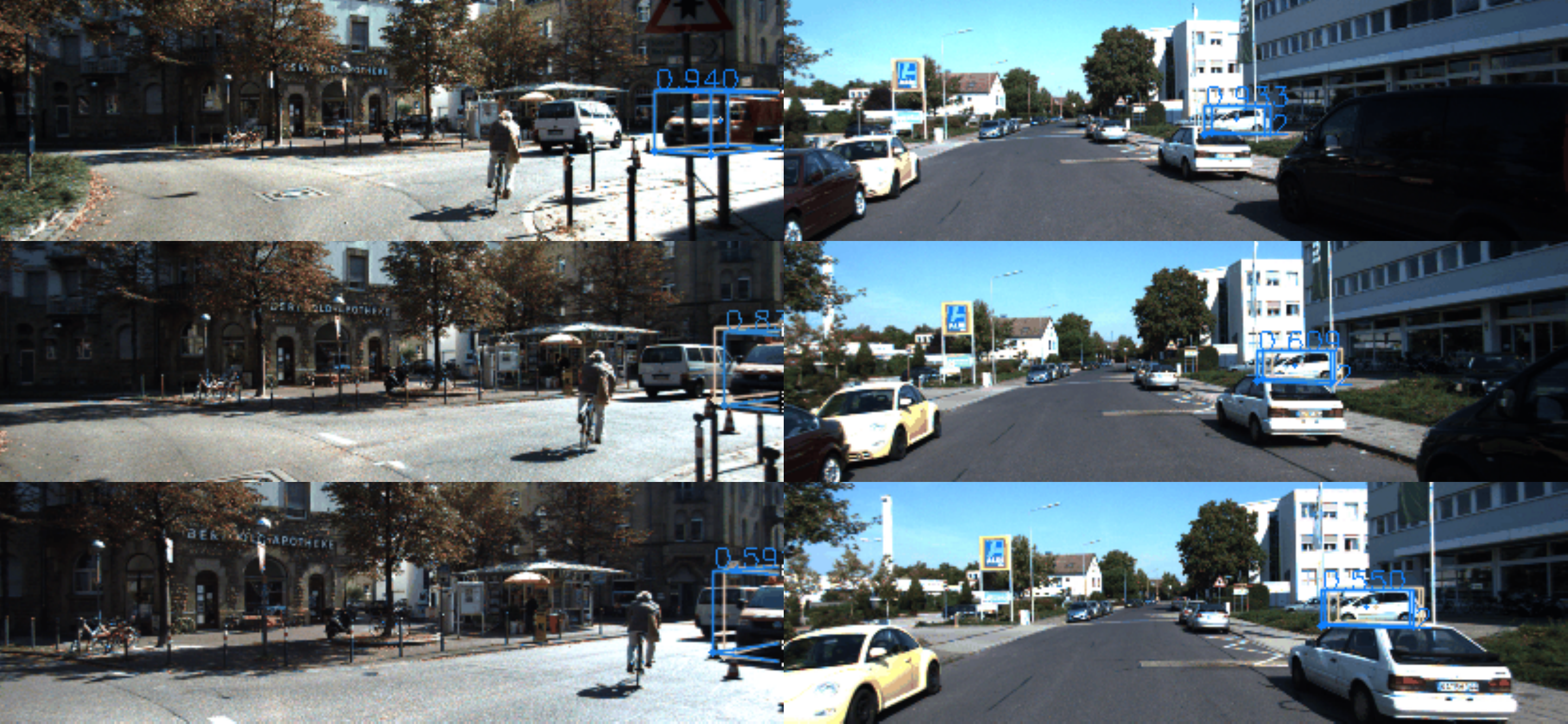}
 \vspace{-0.5em}
 \caption{ 
 \textbf{3D object tracking in KITTI.} 
 IoU scores are marked alongside each estimated box (in blue) across subsampled frames. 
 }
  \vspace{-0.5em}
 \label{fig:tracking_vis}
\end{figure}

For object discovery, We show object proposals of our model in CATER and KITTI in Figure~\ref{fig:dets}. Ground-truth boxes are shown in beige and proposed boxes are shown in blue. Their IoU are marked near the boxes. Results are shown on RGB image as well as bird's-eye view. The boxes have high recall and high precision overall; it can detect small objects as well as separate the object that are spatially close to each other. In KITTI, there are some false positive results on bushes and trees because of the lack of pseudo-label supervision there. 
We visualize KITTI object tracking in Figure~\ref{fig:tracking_vis}, but we encourage the reader to inspect the supplementary video for clearer tracking visualizations\footnote{\url{http://www.cs.cmu.edu/~aharley/em_cvpr21/}}. 
\subsection{Limitations}\label{sec:limitations}

The proposed method has two main limitations. Firstly, our work assumes access to RGB-D data with accurate depth, which excludes the method from application to general videos (e.g., from YouTube). 
Second, it is unclear how best to mine for negatives (i.e., ``not a moving object"). Right now we use a small region around each pseudo label as negative, but it leaves the method prone to false positives in far-away non-objects like bushes and trees.  

\section{Conclusion} \label{sec:conclusion}

We propose an unsupervised method for detecting and tracking objects in unlabelled RGB-D videos. 
We begin with a simple handcrafted technique for segmenting independently-moving objects from the background, relying on cycle-consistent flows and RANSAC. 
We then train an ensemble of 2D and 3D detectors with these segmentations, under heavy data augmentation.
We then use these detectors to re-label the dataset more densely, and return to the training step. The ensemble agreement keeps precision of the pseudo-labels high, and the data augmentations allow recall to gradually expand. 
Our approach opens new avenues for learning object detectors from videos in arbitrary environments, without requiring explicit object supervision. 

\vspace{-0.5em}
\paragraph{Acknowledgements} This material is based upon work supported by US Army contract W911NF20D0002, Sony AI, DARPA Machine Common Sense, an NSF CAREER award, and the Air Force Office of Scientific Research under award number FA9550-20-1-0423. Any opinions, findings and conclusions or recommendations expressed in this material are those of the author(s) and do not necessarily reflect the views of the United States Army or the United States Air Force.

\clearpage


{\small
\begin{thebibliography}{10}\itemsep=-1pt

\bibitem{alexe2010object}
Bogdan Alexe, Thomas Deselaers, and Vittorio Ferrari.
\newblock What is an object?
\newblock In {\em 2010 IEEE computer society conference on computer vision and
  pattern recognition}, pages 73--80. IEEE, 2010.

\bibitem{alexe2012measuring}
Bogdan Alexe, Thomas Deselaers, and Vittorio Ferrari.
\newblock Measuring the objectness of image windows.
\newblock {\em IEEE transactions on pattern analysis and machine intelligence},
  34(11):2189--2202, 2012.

\bibitem{allen2020towards}
Zeyuan Allen-Zhu and Yuanzhi Li.
\newblock Towards understanding ensemble, knowledge distillation and
  self-distillation in deep learning.
\newblock {\em arXiv preprint arXiv:2012.09816}, 2020.

\bibitem{barnes2018driven}
Dan Barnes, Will Maddern, Geoffrey Pascoe, and Ingmar Posner.
\newblock Driven to distraction: Self-supervised distractor learning for robust
  monocular visual odometry in urban environments.
\newblock In {\em 2018 IEEE International Conference on Robotics and Automation
  (ICRA)}, pages 1894--1900. IEEE, 2018.

\bibitem{bertinetto2016fully}
Luca Bertinetto, Jack Valmadre, Joao~F Henriques, Andrea Vedaldi, and Philip~HS
  Torr.
\newblock Fully-convolutional siamese networks for object tracking.
\newblock In {\em European conference on computer vision}, pages 850--865.
  Springer, 2016.

\bibitem{bhat2020learning}
Goutam Bhat, Felix~Järemo Lawin, Martin Danelljan, Andreas Robinson, Michael
  Felsberg, Luc~Van Gool, and Radu Timofte.
\newblock Learning what to learn for video object segmentation, 2020.

\bibitem{springerlink:10.1007/978-3-642-15555-0_21}
Thomas Brox and Jitendra Malik.
\newblock Object segmentation by long term analysis of point trajectories.
\newblock In Kostas Daniilidis, Petros Maragos, and Nikos Paragios, editors,
  {\em ECCV}, pages 282--295, 2010.

\bibitem{brox2010object}
Thomas Brox and Jitendra Malik.
\newblock Object segmentation by long term analysis of point trajectories.
\newblock In {\em European conference on computer vision}, pages 282--295.
  Springer, 2010.

\bibitem{burgess2019monet}
Christopher~P Burgess, Loic Matthey, Nicholas Watters, Rishabh Kabra, Irina
  Higgins, Matt Botvinick, and Alexander Lerchner.
\newblock Monet: Unsupervised scene decomposition and representation.
\newblock {\em arXiv preprint arXiv:1901.11390}, 2019.

\bibitem{10.5555/2898607.2898816}
Andrew Carlson, Justin Betteridge, Bryan Kisiel, Burr Settles, Estevam~R.
  Hruschka, and Tom~M. Mitchell.
\newblock Toward an architecture for never-ending language learning.
\newblock In {\em Proceedings of the Twenty-Fourth AAAI Conference on
  Artificial Intelligence}, AAAI'10, page 1306–1313. AAAI Press, 2010.

\bibitem{chen2020simple}
Ting Chen, Simon Kornblith, Mohammad Norouzi, and Geoffrey Hinton.
\newblock A simple framework for contrastive learning of visual
  representations.
\newblock {\em arXiv preprint arXiv:2002.05709}, 2020.

\bibitem{chen2013neil}
Xinlei Chen, Abhinav Shrivastava, and Abhinav Gupta.
\newblock {NEIL}: {E}xtracting visual knowledge from web data.
\newblock In {\em ICCV}, pages 1409--1416, 2013.

\bibitem{cheng2019improving}
Jiyu Cheng, Yuxiang Sun, and Max Q-H Meng.
\newblock Improving monocular visual slam in dynamic environments: an
  optical-flow-based approach.
\newblock {\em Advanced Robotics}, 33(12):576--589, 2019.

\bibitem{10.1109/ICCV.1995.466815}
J. Costeira and T. Kanade.
\newblock A multi-body factorization method for motion analysis.
\newblock {\em ICCV}, 1995.

\bibitem{craton1996development}
Lincoln~G Craton.
\newblock The development of perceptual completion abilities: Infants'
  perception of stationary, partially occluded objects.
\newblock {\em Child Development}, 67(3):890--904, 1996.

\bibitem{creswell2020alignnet}
Antonia Creswell, Kyriacos Nikiforou, Oriol Vinyals, Andre Saraiva, Rishabh
  Kabra, Loic Matthey, Chris Burgess, Malcolm Reynolds, Richard Tanburn, Marta
  Garnelo, et~al.
\newblock Alignnet: Unsupervised entity alignment.
\newblock {\em arXiv preprint arXiv:2007.08973}, 2020.

\bibitem{dietterich2000ensemble}
Thomas~G Dietterich.
\newblock Ensemble methods in machine learning.
\newblock In {\em International workshop on multiple classifier systems}, pages
  1--15. Springer, 2000.

\bibitem{duan2019centernet}
Kaiwen Duan, Song Bai, Lingxi Xie, Honggang Qi, Qingming Huang, and Qi Tian.
\newblock Centernet: Keypoint triplets for object detection.
\newblock In {\em Proceedings of the IEEE International Conference on Computer
  Vision}, pages 6569--6578, 2019.

\bibitem{NIPS2016_52947e0a}
S.~M.~Ali Eslami, Nicolas Heess, Theophane Weber, Yuval Tassa, David
  Szepesvari, koray kavukcuoglu, and Geoffrey~E Hinton.
\newblock Attend, infer, repeat: Fast scene understanding with generative
  models.
\newblock In D. Lee, M. Sugiyama, U. Luxburg, I. Guyon, and R. Garnett,
  editors, {\em Advances in Neural Information Processing Systems}, volume~29,
  pages 3225--3233. Curran Associates, Inc., 2016.

\bibitem{Fragkiadaki_2015_CVPR}
Katerina Fragkiadaki, Pablo Arbelaez, Panna Felsen, and Jitendra Malik.
\newblock Learning to segment moving objects in videos.
\newblock In {\em CVPR}, June 2015.

\bibitem{Fragkiadaki:topology}
Katerina Fragkiadaki and Jianbo Shi.
\newblock Detection free tracking: Exploiting motion and topology for
  segmenting and tracking under entanglement.
\newblock In {\em CVPR}, 2011.

\bibitem{fragkiadaki2012video}
Katerina Fragkiadaki, Geng Zhang, and Jianbo Shi.
\newblock Video segmentation by tracing discontinuities in a trajectory
  embedding.
\newblock In {\em 2012 IEEE Conference on Computer Vision and Pattern
  Recognition}, pages 1846--1853. IEEE, 2012.

\bibitem{Geiger2013IJRR}
Andreas Geiger, Philip Lenz, Christoph Stiller, and Raquel Urtasun.
\newblock Vision meets robotics: The kitti dataset.
\newblock {\em International Journal of Robotics Research (IJRR)}, 2013.

\bibitem{girdhar2020cater}
Rohit Girdhar and Deva Ramanan.
\newblock {CATER: A diagnostic dataset for Compositional Actions and TEmporal
  Reasoning}.
\newblock In {\em ICLR}, 2020.

\bibitem{goldstein2009perceiving}
E~Bruce Goldstein.
\newblock Perceiving objects and scenes: the gestalt approach to object
  perception.
\newblock {\em Goldstein EB. Sensation and Perception. 8th ed. Belmont, CA:
  Wadsworth Cengage Learning}, 2009.

\bibitem{greff2019multi}
Klaus Greff, Rapha{\"e}l~Lopez Kaufman, Rishabh Kabra, Nick Watters,
  Christopher Burgess, Daniel Zoran, Loic Matthey, Matthew Botvinick, and
  Alexander Lerchner.
\newblock Multi-object representation learning with iterative variational
  inference.
\newblock In {\em International Conference on Machine Learning}, pages
  2424--2433. PMLR, 2019.

\bibitem{tracking_emerges_3d}
Adam~W Harley, Shrinidhi~K Lakshmikanth, Paul Schydlo, and Katerina
  Fragkiadaki.
\newblock Tracking emerges by looking around static scenes, with neural {3D}
  mapping.
\newblock In {\em ECCV}, 2020.

\bibitem{he2019momentum}
Kaiming He, Haoqi Fan, Yuxin Wu, Saining Xie, and Ross Girshick.
\newblock Momentum contrast for unsupervised visual representation learning.
\newblock In {\em CVPR}, 2020.

\bibitem{he2016deep}
Kaiming He, Xiangyu Zhang, Shaoqing Ren, and Jian Sun.
\newblock Deep residual learning for image recognition.
\newblock In {\em Proceedings of the IEEE conference on computer vision and
  pattern recognition}, pages 770--778, 2016.

\bibitem{hinton2015distilling}
Geoffrey Hinton, Oriol Vinyals, and Jeff Dean.
\newblock Distilling the knowledge in a neural network.
\newblock {\em arXiv preprint arXiv:1503.02531}, 2015.

\bibitem{Hu_2019_CVPR}
Yuan-Ting Hu, Hong-Shuo Chen, Kexin Hui, Jia-Bin Huang, and Alexander~G.
  Schwing.
\newblock Sail-vos: Semantic amodal instance level video object segmentation -
  a synthetic dataset and baselines.
\newblock In {\em The IEEE Conference on Computer Vision and Pattern
  Recognition (CVPR)}, June 2019.

\bibitem{itti2000saliency}
Laurent Itti and Christof Koch.
\newblock A saliency-based search mechanism for overt and covert shifts of
  visual attention.
\newblock {\em Vision research}, 40(10-12):1489--1506, 2000.

\bibitem{Jiang*2020SCALOR:}
Jindong Jiang*, Sepehr Janghorbani*, Gerard~De Melo, and Sungjin Ahn.
\newblock Scalor: Generative world models with scalable object representations.
\newblock In {\em International Conference on Learning Representations}, 2020.

\bibitem{johnson2016clevr}
Justin Johnson, Bharath Hariharan, Laurens van~der Maaten, Li Fei-Fei,
  C.~Lawrence Zitnick, and Ross Girshick.
\newblock Clevr: A diagnostic dataset for compositional language and elementary
  visual reasoning, 2016.

\bibitem{keller2013real}
Maik Keller, Damien Lefloch, Martin Lambers, Shahram Izadi, Tim Weyrich, and
  Andreas Kolb.
\newblock Real-time 3d reconstruction in dynamic scenes using point-based
  fusion.
\newblock In {\em 2013 International Conference on 3D Vision-3DV 2013}, pages
  1--8. IEEE, 2013.

\bibitem{kerl13iros}
C. Kerl, J. Sturm, and D. Cremers.
\newblock Dense visual {SLAM} for {RGB-D} cameras.
\newblock In {\em IROS}, 2013.

\bibitem{DBLP:journals/corr/KhorevaBIBS17}
Anna Khoreva, Rodrigo Benenson, Eddy Ilg, Thomas Brox, and Bernt Schiele.
\newblock Lucid data dreaming for object tracking.
\newblock In {\em CVPR Workshops}, 2017.

\bibitem{kingma2014adam}
Diederik~P Kingma and Jimmy Ba.
\newblock {Adam}: {A} method for stochastic optimization.
\newblock {\em arXiv preprint arXiv:1412.6980}, 2014.

\bibitem{kulkarni2019unsupervised}
Tejas Kulkarni, Ankush Gupta, Catalin Ionescu, Sebastian Borgeaud, Malcolm
  Reynolds, Andrew Zisserman, and Volodymyr Mnih.
\newblock Unsupervised learning of object keypoints for perception and control.
\newblock In {\em NeurIPS}, 2019.

\bibitem{li2020fast}
Yu Li, Zhuoran Shen, and Ying Shan.
\newblock Fast video object segmentation using the global context module, 2020.

\bibitem{lin2017feature}
Tsung-Yi Lin, Piotr Doll{\'a}r, Ross Girshick, Kaiming He, Bharath Hariharan,
  and Serge Belongie.
\newblock Feature pyramid networks for object detection.
\newblock In {\em Proceedings of the IEEE conference on computer vision and
  pattern recognition}, pages 2117--2125, 2017.

\bibitem{selflow}
Pengpeng Liu, Michael Lyu, Irwin King, and Jia Xu.
\newblock Selflow: Self-supervised learning of optical flow.
\newblock In {\em Proceedings of the IEEE Conference on Computer Vision and
  Pattern Recognition}, pages 4571--4580, 2019.

\bibitem{locatello2020object}
Francesco Locatello, Dirk Weissenborn, Thomas Unterthiner, Aravindh Mahendran,
  Georg Heigold, Jakob Uszkoreit, Alexey Dosovitskiy, and Thomas Kipf.
\newblock Object-centric learning with slot attention.
\newblock {\em Advances in Neural Information Processing Systems}, 33, 2020.

\bibitem{matthews2004template}
Lain Matthews, Takahiro Ishikawa, and Simon Baker.
\newblock The template update problem.
\newblock {\em IEEE transactions on pattern analysis and machine intelligence},
  26(6):810--815, 2004.

\bibitem{flyingthings16}
N. Mayer, E. Ilg, P. H\"ausser, P. Fischer, D. Cremers, A. Dosovitskiy, and T.
  Brox.
\newblock A large dataset to train convolutional networks for disparity,
  optical flow, and scene flow estimation.
\newblock In {\em CVPR}, 2016.

\bibitem{mitchell2018never}
Tom Mitchell, William Cohen, Estevam Hruschka, Partha Talukdar, Bishan Yang,
  Justin Betteridge, Andrew Carlson, Bhavana Dalvi, Matt Gardner, Bryan Kisiel,
  et~al.
\newblock Never-ending learning.
\newblock {\em Communications of the ACM}, 61(5):103--115, 2018.

\bibitem{nigam2000text}
Kamal Nigam, Andrew~Kachites McCallum, Sebastian Thrun, and Tom Mitchell.
\newblock Text classification from labeled and unlabeled documents using {EM}.
\newblock {\em Machine learning}, 39(2):103--134, 2000.

\bibitem{OB11}
P. Ochs and T. Brox.
\newblock Object segmentation in video: {A} hierarchical variational approach
  for turning point trajectories into dense regions.
\newblock In {\em ICCV}, 2011.

\bibitem{pan2009recurrent}
Pan Pan, Fatih Porikli, and Dan Schonfeld.
\newblock Recurrent tracking using multifold consistency.
\newblock In {\em Proceedings of the Eleventh IEEE International Workshop on
  Performance Evaluation of Tracking and Surveillance}, 2009.

\bibitem{8099855}
F. {Perazzi}, A. {Khoreva}, R. {Benenson}, B. {Schiele}, and A.
  {Sorkine-Hornung}.
\newblock Learning video object segmentation from static images.
\newblock In {\em 2017 IEEE Conference on Computer Vision and Pattern
  Recognition (CVPR)}, pages 3491--3500, 2017.

\bibitem{Radosavovic_2018_CVPR}
Ilija Radosavovic, Piotr Dollár, Ross Girshick, Georgia Gkioxari, and Kaiming
  He.
\newblock Data distillation: Towards omni-supervised learning.
\newblock In {\em Proceedings of the IEEE Conference on Computer Vision and
  Pattern Recognition (CVPR)}, June 2018.

\bibitem{ronneberger2015u}
Olaf Ronneberger, Philipp Fischer, and Thomas Brox.
\newblock U-net: Convolutional networks for biomedical image segmentation.
\newblock In {\em International Conference on Medical image computing and
  computer-assisted intervention}, pages 234--241. Springer, 2015.

\bibitem{sarkar1993perceptual}
Sudeep Sarkar and Kim~L Boyer.
\newblock Perceptual organization in computer vision: A review and a proposal
  for a classificatory structure.
\newblock {\em IEEE Transactions on Systems, Man, and Cybernetics},
  23(2):382--399, 1993.

\bibitem{schoeps14ismar}
T. Sch\"ops, J. Engel, and D. Cremers.
\newblock Semi-dense visual odometry for {AR} on a smartphone.
\newblock In {\em ISMAR}, 2014.

\bibitem{sethi1987finding}
Ishwar~K Sethi and Ramesh Jain.
\newblock Finding trajectories of feature points in a monocular image sequence.
\newblock {\em IEEE Transactions on pattern analysis and machine intelligence},
  9(1):56--73, 1987.

\bibitem{spelke1982perceptual}
Elizabeth~S Spelke, J Mehler, M Garrett, and E Walker.
\newblock Perceptual knowledge of objects in infancy.
\newblock In NJ:~Erlbaum Hillsdale, editor, {\em Perspectives on mental
  representation}, chapter~22. Erlbaum, 1982.

\bibitem{raftnet}
Zachary Teed and Jia Deng.
\newblock Raft: Recurrent all-pairs field transforms for optical flow.
\newblock In {\em European Conference on Computer Vision}, pages 402--419.
  Springer, 2020.

\bibitem{tobin2017domain}
Josh Tobin, Rachel Fong, Alex Ray, Jonas Schneider, Wojciech Zaremba, and
  Pieter Abbeel.
\newblock Domain randomization for transferring deep neural networks from
  simulation to the real world.
\newblock In {\em 2017 IEEE/RSJ international conference on intelligent robots
  and systems (IROS)}, pages 23--30. IEEE, 2017.

\bibitem{Tomasi:1992:SMI:144398.144403}
Carlo Tomasi and Takeo Kanade.
\newblock Shape and motion from image streams under orthography: A
  factorization method.
\newblock {\em Int. J. Comput. Vision}, 9(2):137--154, Nov. 1992.

\bibitem{vondrick2018tracking}
Carl Vondrick, Abhinav Shrivastava, Alireza Fathi, Sergio Guadarrama, and Kevin
  Murphy.
\newblock Tracking emerges by colorizing videos.
\newblock In {\em Proceedings of the European Conference on Computer Vision
  (ECCV)}, pages 391--408, 2018.

\bibitem{Weng2020_AB3DMOT_eccvw}
Xinshuo Weng, Jianren Wang, David Held, and Kris Kitani.
\newblock {AB3DMOT: A Baseline for 3D Multi-Object Tracking and New Evaluation
  Metrics}.
\newblock {\em ECCVW}, 2020.

\bibitem{wu2007situ}
Hao Wu, Aswin~C Sankaranarayanan, and Rama Chellappa.
\newblock In situ evaluation of tracking algorithms using time reversed chains.
\newblock In {\em 2007 IEEE Conference on Computer Vision and Pattern
  Recognition}, pages 1--8. IEEE, 2007.

\bibitem{yang2020teaser}
Heng Yang, Jingnan Shi, and Luca Carlone.
\newblock {TEASER}: {F}ast and certifiable point cloud registration.
\newblock {\em IEEE Transactions on Robotics}, 2020.

\bibitem{back_to_basics:2016}
Jason~J. Yu, Adam~W. Harley, and Konstantinos~G. Derpanis.
\newblock Back to basics: Unsupervised learning of optical flow via brightness
  constancy and motion smoothness.
\newblock In {\em ECCV}, 2016.

\bibitem{zeng2019end}
Wenyuan Zeng, Wenjie Luo, Simon Suo, Abbas Sadat, Bin Yang, Sergio Casas, and
  Raquel Urtasun.
\newblock End-to-end interpretable neural motion planner.
\newblock In {\em Proceedings of the IEEE Conference on Computer Vision and
  Pattern Recognition}, pages 8660--8669, 2019.

\end{thebibliography}

}

\appendix

\section*{Appendix}
In this appendix, we provide implementation details for the self-supervision (Sec.~\ref{sec:selfsupimpl}) and training (Sec.~\ref{sec:trainingimpl}), and some additional analysis of the baselines' object discovery performance (Sec.~\ref{sec:objdiscoveryanalysis}). We also encourage the reader to watch our supplementary video (\url{http://www.cs.cmu.edu/~aharley/em_cvpr21/}), which illustrates our incremental pseudolabel generation process for a sample video sequence from KITTI, and shows qualitative tracking results in both KITTI and CATER for our method and the baselines. 


\section{Self-supervision details}\label{sec:selfsupimpl}

\begin{figure}[t!]
\centering{
 \includegraphics[width=1.0\linewidth]{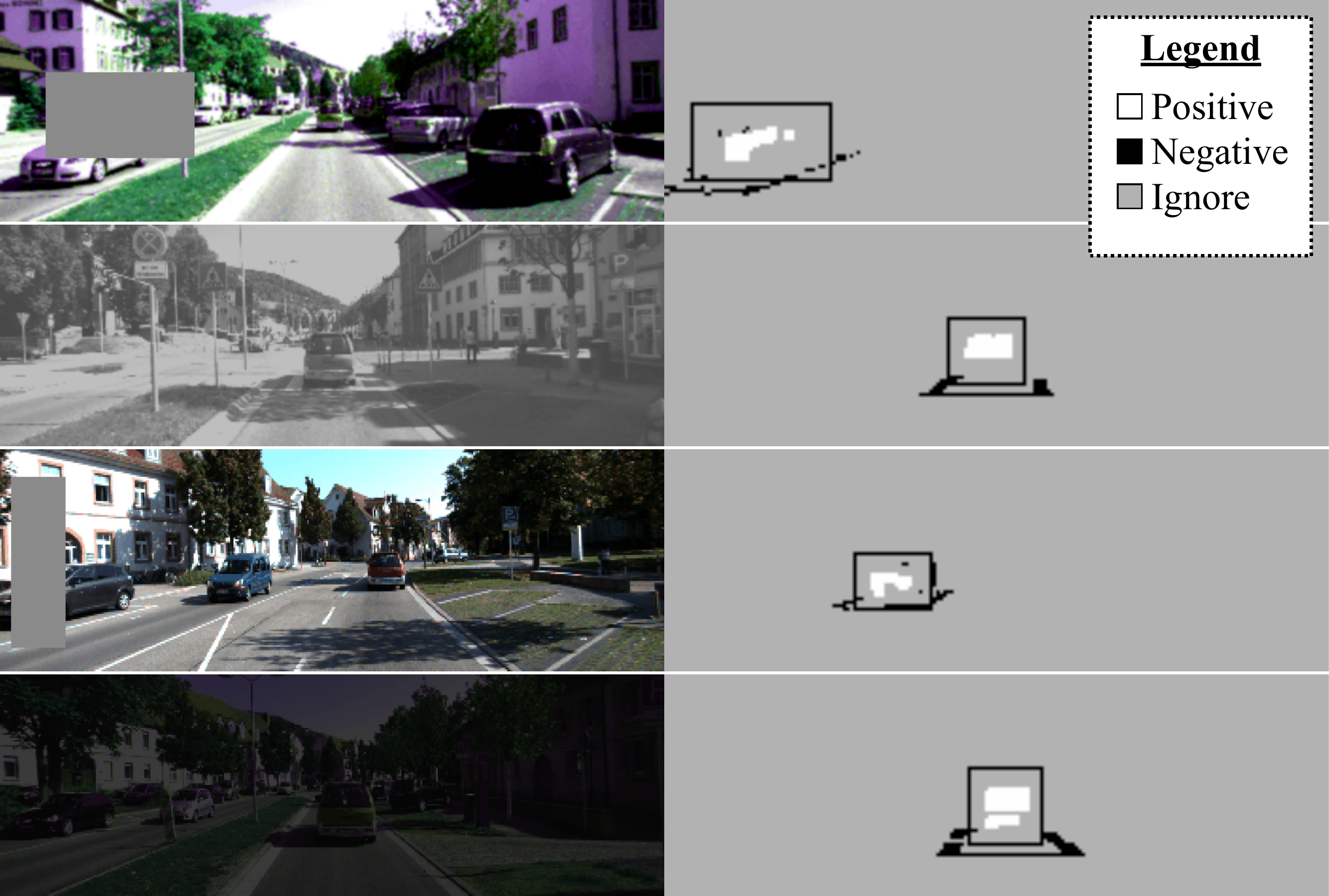}
 }
 \caption{ 
 \textbf{Sparse supervision for 2D objectness segmentation.} Left: RGB images with color and occlusion augmentations. Right: supervision generated from pseudolabels. Supervision is only generated in the region immediately surrounding the discovered objects.}
 \label{fig:label2d}
\end{figure}

After converting estimates into pseudolabels, it is important to design the supervision strategy so that the appearance-based objectness modules have the potential to generalize from this data to the yet-unannotated data. This involves (1) heavy augmentations, and (2) converting the pseudolabels into supervision regions:
\begin{itemize}
    \item \textbf{Positive}: regions where we have detected a moving object;
    \item \textbf{Negative}: regions where a moving object is unlikely;
    \item \textbf{Ignore}: regions where the pseudolabels are ambiguous about moving/non-moving.
\end{itemize}
The ``ignore'' region is critical: If we simply used the pseudolabels as positives and treated the remainder of the scene as negatives, the modules would effectively be trained to \textit{not} detect objects beyond the annotated ones. The design of the regions is slightly different for the 2D segmentor vs. the 3D object detector. 

For 2D objectness segmentation, we need to annotate pixels. To create positive pixels, we shrink the estimated bounding boxes (by 0.1m on each side), then collect pointcloud points that are within those boxes, and project those points into the image. To create negative pixels, we enlarge the boxes (by 2.0m on each side), and collect points that are between the original-sized box and the enlarged box, and project those into the image. Additionally, we project the box itself into the image, and use its outer contour as negative pixels. We leave all other pixels as ``ignore'', and do not apply loss there. These supervision labels are illustrated in Figure~\ref{fig:label2d}.

For 3D detection, we annotate voxels on a 3D grid. Since we use a CenterNet-style detector \cite{duan2019centernet}, we do not need to annotate anchor boxes as positives or negatives, but instead create a ``target'' objectness heatmap (indicating positives and negatives densely), with a small Gaussian at the centroid of each annotated object. At the peak of the Gaussian, we supply the subvoxel offset and orientation information, in the additional channels. 
As in the original CenterNet, we only supervise offset and orientation at the centroid itself. To avoid penalizing detections in the unannotated part of the scene, we treat all voxels that lie beyond a radius of an annotation as part of the ``ignore'' region, and do not apply loss there. We set the radius according to the annotation, using a value of $r = 3 \cdot \max(l, h, w)$, where $l, h, w$ denote the length, height, and width of the annotated object.

Within each batch, we evenly balance the loss induced by positive labels and negative labels, to ensure that the model does not learn a frequency bias for either class. 

\section{Training details}\label{sec:trainingimpl}
We optimize the parameters of all CNNs with the Adam optimizer \cite{kingma2014adam}, with an effective batch size of 4. To achieve this on smaller GPUs with memory constraints, we use a batch size of 1 but accumulate gradients from 4 steps before taking a step with the optimizer.

For the 2D segmentor and the 3D object detector, in the first training round, we use random initialization of the networks, and set the learning rate to $1e-4$. In the second training round, we use the first round's parameters as initialization, and train with a learning rate of $1e-5$. The segmentor converges in approximately 40,000 iterations, while the 3D detector requires approximately 80,000 iterations. For the optical flow module, we initialize with the RAFT model trained on FlyingThings \cite{flyingthings16}. We have also fine-tuned the RAFT model with standard self-supervision objectives (brightness constancy, smoothness, forward-backward consistency), without significant improvement beyond this strong initialization. 

In CATER we use an image resolution of $128 \times 384$, and a 3D voxel grid resolution of $128 \times 64 \times 128$, spanning a range of $8 \times 4 \times 8$ in the dataset's units. In KITTI we use an image resolution of $128 \times 416$, and a 3D voxel grid resolution of $256 \times 32 \times 256$, spanning a metric range of $32m \times 8m \times 32m$. 



\begin{figure}[t]
\centering{
 \includegraphics[width=0.9\linewidth]{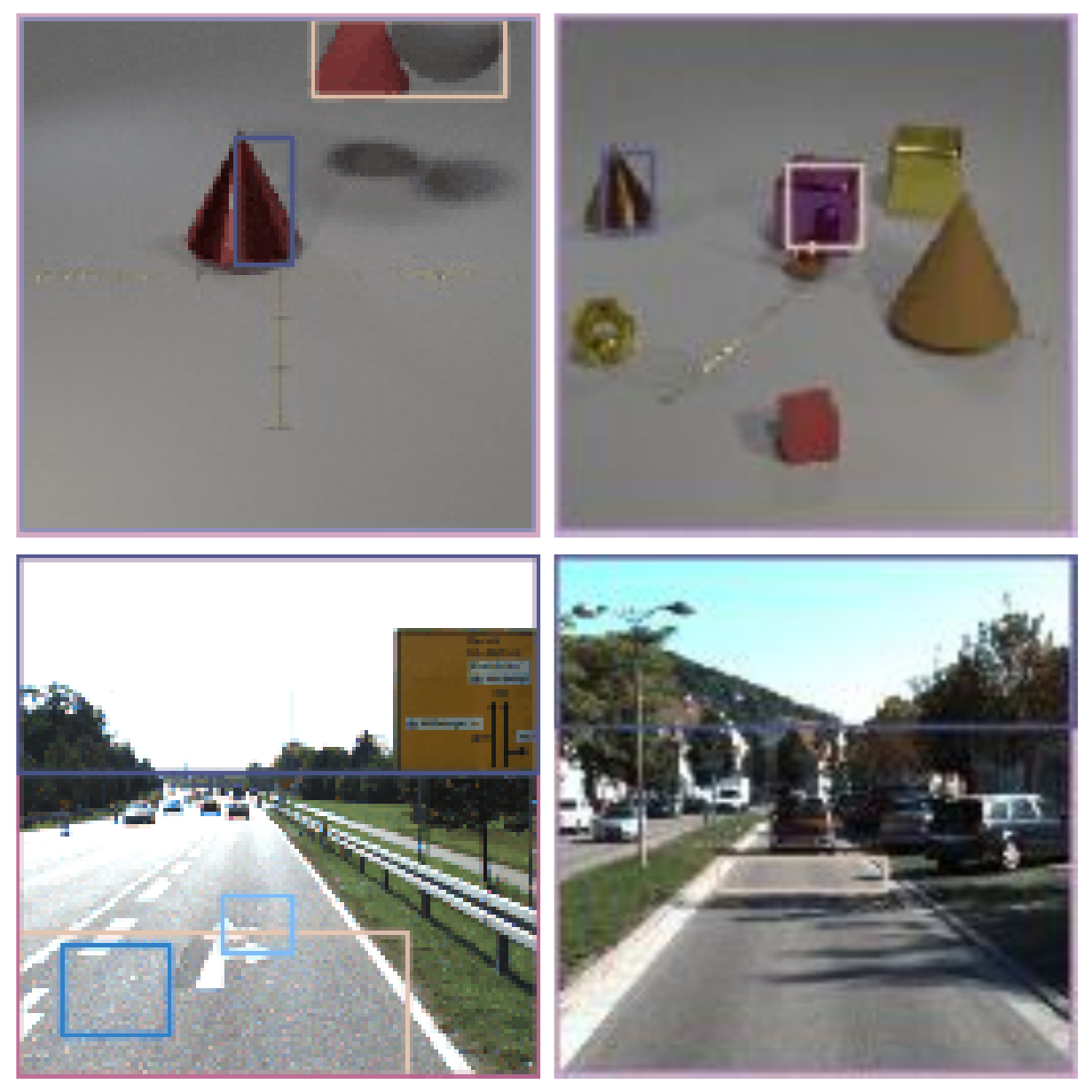}
 }
 \caption{ 
 \textbf{Object discovery results by MONet \cite{burgess2019monet} in CATER (top) and KITTI (bottom).} In CATER the proposed boxes are typically on objects, but in KITTI we find that the model produces boxes for non-object scene elements, such as segments of the road, lane markings, and the sky.}
 \label{fig:monet_discovery}
\end{figure}

\begin{figure}[t]
\centering{
 \subfigure[]{\includegraphics[width=\linewidth]{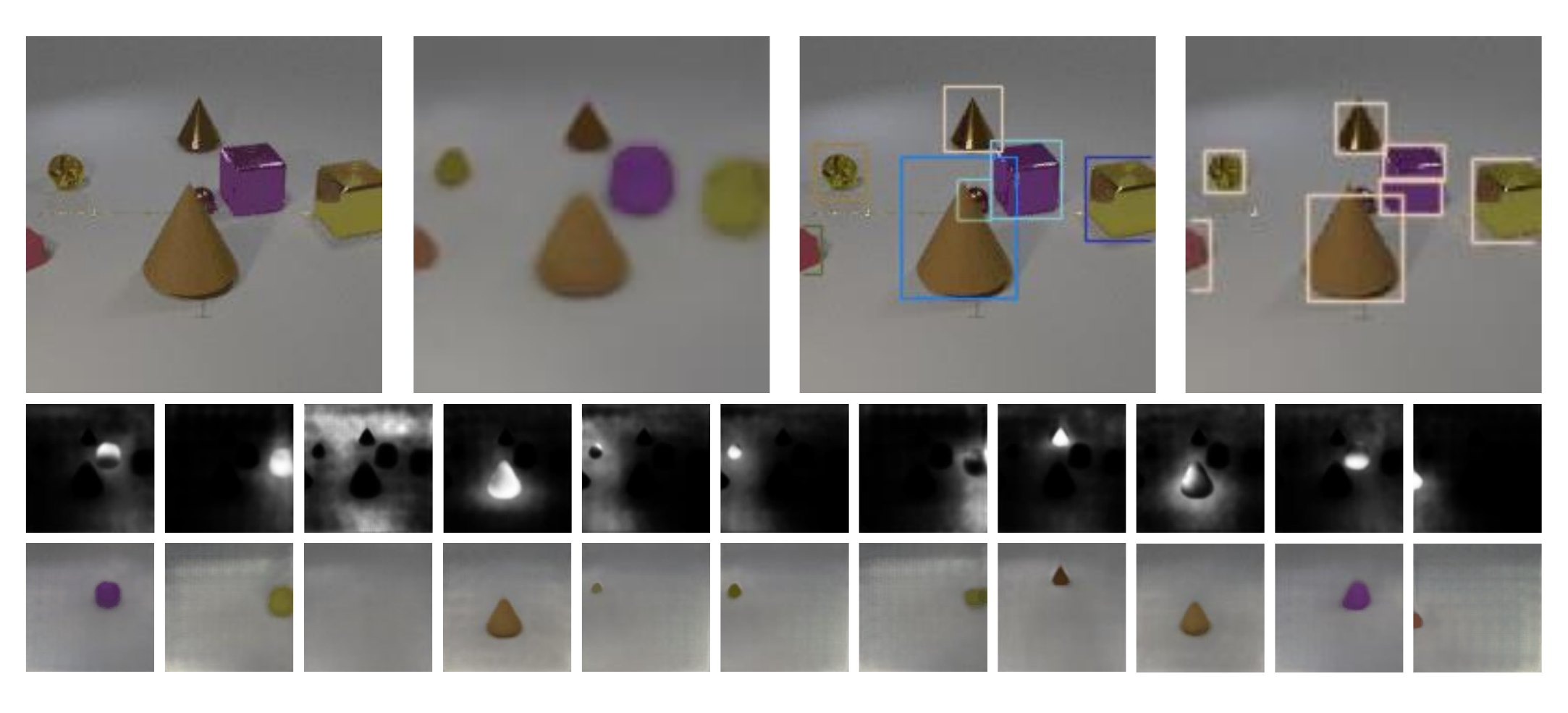}}
 \subfigure[]{\includegraphics[width=\linewidth]{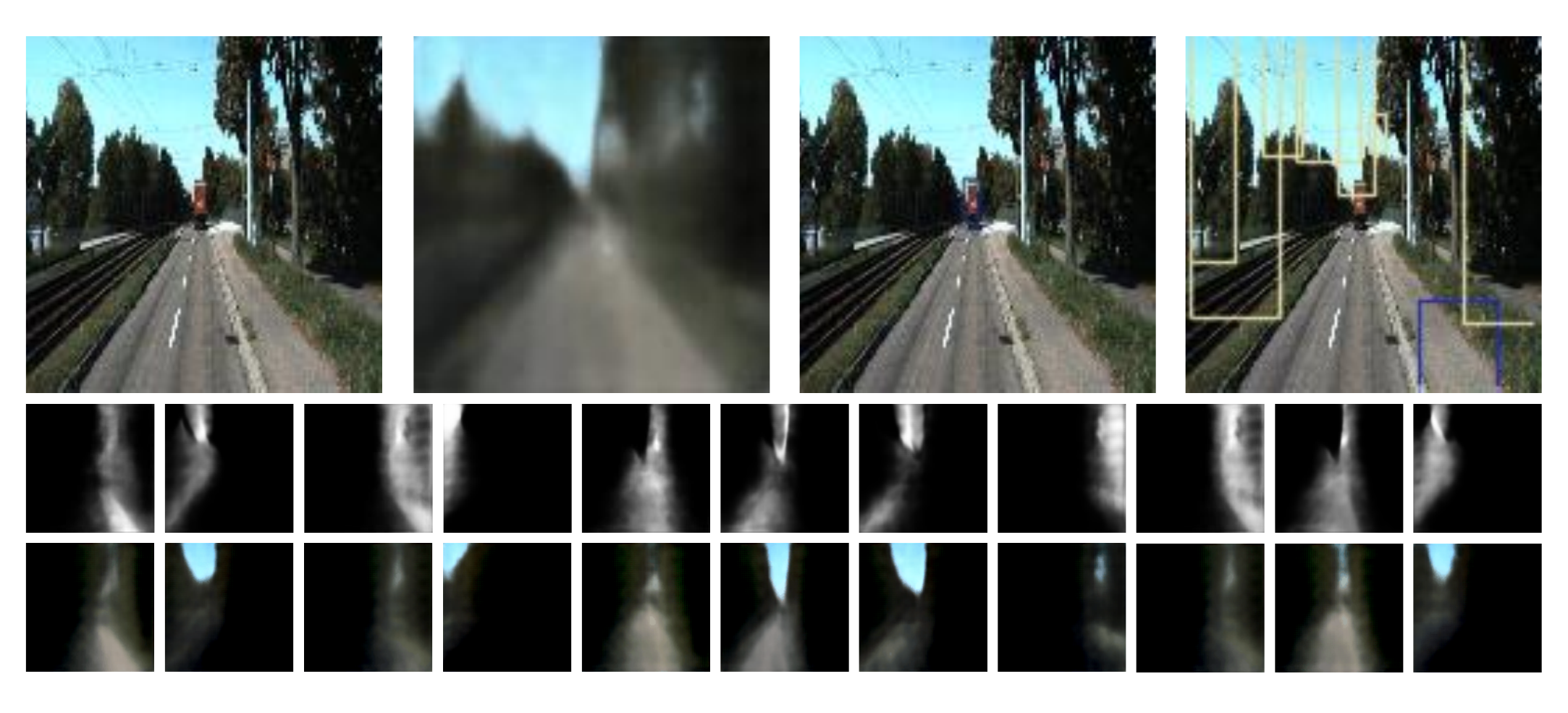}}
 }
 \caption{ 
 \textbf{Object discovery results by Slot Attention \cite{locatello2020object} in CATER (a) and KITTI (b).} For each dataset, the rows are arranged as follows: \textbf{1st row (left to right):} RGB ground truth, RGB reconstruction by the model, ground truth bounding boxes, predicted bounding boxes. \textbf{2nd row:} The predicted masks for each slot. \textbf{3rd row:} RGB reconstruction for each slot. }
 \label{fig:slot_discovery}
\end{figure}

\section{Object discovery analysis}\label{sec:objdiscoveryanalysis}

In the main paper we noted that the baseline methods for object discovery perform reasonably-well in CATER, but fail to produce meaningful outputs in KITTI. We illustrate this in Figure~\ref{fig:monet_discovery} for MONet \cite{burgess2019monet}, and in Figure~\ref{fig:slot_discovery} for Slot Attention \cite{locatello2020object}. We find that optimizing these models is considerably more difficult in a real-world dataset such as KITTI, than it is in simple synthetic datasets. It is likely that some underlying assumptions of these models are violated in the real world, such as: lack of camera motion, within-object self-similarity, and background self-similarity, which would typically allow less-ambiguous figure-ground segmentation.  

\end{document}